\documentclass[conference]{IEEEtran}
\IEEEoverridecommandlockouts
\usepackage{cite}
\usepackage{amsmath,amssymb,amsfonts}
\usepackage{algorithmic}
\usepackage{graphicx}
\usepackage{textcomp}
\usepackage{xcolor}

\usepackage{amsmath}

\DeclareMathOperator*{\argmin}{arg\,min}

\usepackage[linesnumbered,ruled,vlined]{algorithm2e}

\usepackage[symbol]{footmisc}

\usepackage{graphicx}
\ifCLASSOPTIONcompsoc
    \usepackage[caption=false, font=normalsize, labelfont=sf, textfont=sf]{subfig}
\else
\usepackage[caption=false, font=footnotesize]{subfig}
\fi

\usepackage{etoolbox}
\def\tablename{Table}
\makeatletter
\patchcmd{\@makecaption}
  {\scshape}
  {}
  {}
  {}
\patchcmd{\@makecaption}
  {\\}
  {.\ }
  {}
  {}
\makeatother

\usepackage{url}

\def\BibTeX{{\rm B\kern-.05em{\sc i\kern-.025em b}\kern-.08em
    T\kern-.1667em\lower.7ex\hbox{E}\kern-.125emX}}
\begin{document}

\title{Counterfactual Explanations for Time Series Forecasting\\
\thanks{This work was funded in part by the Digital Futures EXTREMUM project.}
}


\author{
\IEEEauthorblockN{Zhendong Wang, Ioanna Miliou, Isak Samsten, and Panagiotis Papapetrou}
\IEEEauthorblockA{\textit{Department of Computer and Systems Sciences} \\
\textit{Stockholm University}, 
Stockholm, Sweden \\
\{zhendong.wang, ioanna.miliou, samsten, panagiotis\}@dsv.su.se}
}

\maketitle

\begin{abstract}
Among recent developments in time series forecasting methods, deep forecasting models have gained popularity as they can utilize hidden feature patterns in time series to improve forecasting performance. 
Nevertheless, the majority of current deep forecasting models are opaque, hence making it challenging to interpret the results. 
While counterfactual explanations have been extensively employed as a post-hoc approach for explaining classification models, their application to forecasting models still remains underexplored.
In this paper, we formulate the novel problem of counterfactual generation for time series forecasting, and propose an algorithm, called ForecastCF, that solves the problem by applying gradient-based perturbations to the original time series. ForecastCF guides the perturbations by applying constraints to the forecasted values to obtain desired prediction outcomes. 
We experimentally evaluate ForecastCF using four state-of-the-art deep model architectures and compare to two baselines. Our results show that ForecastCF outperforms the baseline in terms of counterfactual validity and data manifold closeness. 
Overall, our findings suggest that ForecastCF can generate meaningful and relevant counterfactual explanations for various forecasting tasks.
\end{abstract}


\section{Introduction}
Time series forecasting refers to the process of predicting the future values of an input time series given some past observations and potentially a set of exogenous variables. Statistical and machine learning (ML) models have been developed for time series forecasting, including some recent deep learning (DL) solutions \cite{lim_time-series_2021}. Time series forecasting is highly applicable to several domains and scenarios. 
For example, in the retail industry, a high-quality forecasting model is able to use historical sales data to predict future demand accurately, and can potentially help retailers to plan optimal assortments and make effective daily operational decisions \cite{punia_predictive_2022,makridakis_m5_2022}. 
Time series forecasting has also demonstrated its applicability in the finance domain to support automatic electronic trading systems like in stock and foreign exchange forecasting \cite{sezer_financial_2020}. 
Moreover, in healthcare, forecasting models could help clinical practitioners with prognostic tasks to understand how the health status of a patient is evolving \cite{vo_van_time_2018,song_attend_2018}; but also, it could support hospitals in planning the amount of newly admitted patients and arrange the resources in an effortless manner \cite{zhou_time_2018}.

The advantage of deep forecasting models is that they can potentially utilize hidden feature patterns in time series to improve the forecasting performance compared to classical statistical models (e.g., exponential smoothing and Autoregressive Integrated Moving Average (ARIMA)) \cite{bandara_forecasting_2020,petropoulos_forecasting_2022}. Most classical statistical models are designed for single univariate time series, which may fail to produce reliable forecasts because they consider each individual time series in isolation during modeling \cite{hewamalage_recurrent_2021}.
However, in real-world scenarios, typically, there are multiple similar time series rather than a single individual series of interest. For example, in predictive maintenance and electric load forecasting, it has been proven more effective to apply a global deep forecasting model to learn from large amounts of similar time series \cite{bandara_forecasting_2020,toubeau_deep_2019}. At the same time, the Amazon research team also claimed that it is more beneficial to deploy deep models for operational forecasting in terms of multiple product sales \cite{faloutsos_classical_2019}. 

Earlier works have proposed different neural network architectures to address forecasting tasks, ranging from Recurrent Neural Networks (RNN) \cite{lim_time-series_2021} to more advanced architectures, like Transformers \cite{li_enhancing_2019}. 
Moreover, in recent forecast competitions, deep forecasting models based on these architectures have demonstrated competitive results. For example, the M4 competition winner was an RNN-based model with exponential smoothing \cite{smyl_hybrid_2020}; while the DeepAR model, also using RNN architectures, won the third place in the M5 competition \cite{salinas_deepar_2020}. 
However, the majority of the current deep forecasting models are considered ``black-box'' models due to the non-linearity in the network architectures, and it remains challenging to interpret the forecasting results. 

Understanding the most important factors that contribute to the forecasting outcome is crucial for the end-users to trust the predictions and hence improve the model quality \cite{kamath_explainability_2021}. 
In order to provide model explainability, one can apply post-hoc methods like LIME \cite{ribeiro_why_2016} or SHAP \cite{lundberg_unified_2017} to explain prediction outcomes; specifically, for time series forecasting, they have been adopted to provide feature importance scores from the input space \cite{schlegel_ts-mule_2021,zhang_fi-shap_2022}. 
Additionally, saliency maps and association rules have been explored to provide visual explanations for deep forecasting models \cite{saadallah_explainable_2021,troncoso-garcia_new_2023}. 
Nonetheless, recent research focused mostly on highlighting the important segments of the time series for deep forecasting models; little emphasis has been given on actionability, and how can a forecasting outcome be changed from an undesired state to a desired one.

Counterfactual explanations were initially proposed by Wachter et el. \cite{wachter_counterfactual_2017} as a post-hoc explanation approach for ML classification models, which intend to modify input data features to perturb the target label to a desired state. 
Specifically, for time series classification, counterfactuals have been applied to explain which part of the input time series could be modified such that the classifier predicts the desired target label. Recent research demonstrated that counterfactuals are applicable both in univariate and multivariate time series classification problems\cite{karlsson_explainable_2018,ates_counterfactual_2021,wang_learning_2021,delaney_instance-based_2021}. 
The advantage of counterfactual explanations is that counterfactuals are model-agnostic and do not require additional information (or modifications) from the black-box models, except the prediction function. 
However, to the best of our knowledge, counterfactuals have not been defined for the task of time series forecasting over a forecasting time horizon. Defining time series counterfactuals for forecasting can be beneficial in several application areas as it can provide potentially actionable interventions for altering the forecasting outcome to, e.g., a more favorable trend, within the time horizon.

\begin{figure}[t]
\centering
\includegraphics[width=0.8\linewidth]{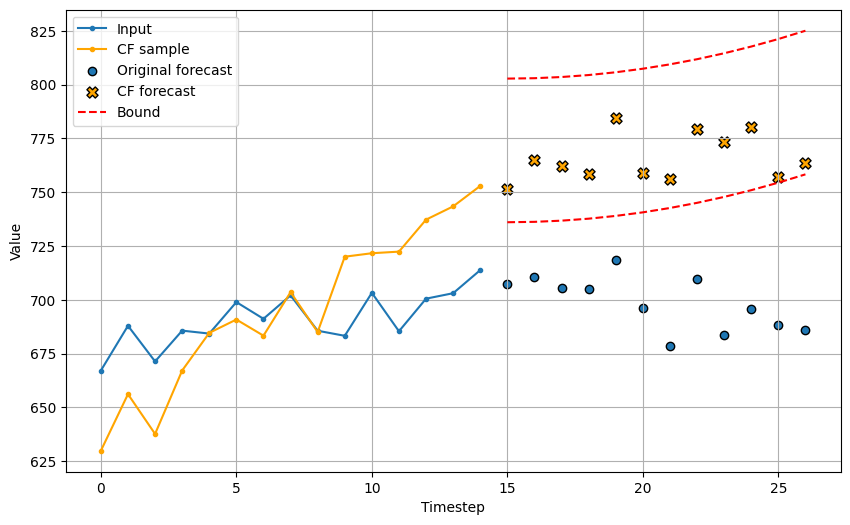}
\caption{Example of a time series counterfactual (yellow line) of an input time series (blue line) from the finance sector, where, given a prediction constraint (red-dotted lines), we would like the underlying forecasting model to predict values (yellow x-points) that fall within that band.}
\label{fig:example}
\vspace{-4mm}
\end{figure}

\subsection{Example}
In the example depicted in Figure \ref{fig:example}, we can observe the number of sales of a product over the last 14 days (blue line). Using a state-of-the-art time series forecasting model, we can predict the number of sales for the next 12 days (blue dots). Suppose now that our objective is to affect the expected future values so that they increase substantially and keep following an upward-growing trend (red-dotted lines). We should then generate a counterfactual of the sales over the past 14 days (yellow line) so that our forecasting model predicts the future sales values (yellow x-points) to be within the desired forecasting trend and value bounds. Our proposed counterfactual is actionable by suggesting value and trend changes in the past that will, in turn, result in aligning the future sales trend to the desired sales forecasting trend. Such changes in the past can be imposed by external interventions, such as marketing campaigns that will boost sales (days 9-14), or pricing policies that will reduce the sales by increasing the product price (days 0-6). 

\subsection{Contributions}
Our main contributions in this paper can be summarized as follows:
\begin{itemize}
\item We formulate the novel problem of counterfactual explanations for time series forecasting, and demonstrate its applicability to several application domains.
\item We propose a gradient-based algorithm for generating time series counterfactuals so that the forecasted values over a time horizon satisfy a set of lower and upper bound constraints.
\item We perform an extensive experimental evaluation on six datasets and evaluate ForecastCF on four DL model architectures, including state-of-the-art models for univariate time series forecasting.
\end{itemize}

\section{Background}
\subsection{Related Work}
\noindent \textbf{Forecasting with deep learning}:
Time series forecasting has been well studied in the research community for decades, including classical univariate models like ARIMA and exponential smoothing models \cite{petropoulos_forecasting_2022}.
More recently, a variety of RNN-based models have been studied, e.g., Gated Recurrent Unit (GRU) and Long Short-Term Memory (LSTM) variants \cite{lim_time-series_2021,hewamalage_recurrent_2021,bianchi_overview_2017}. In addition, different techniques like skip-connections and attention weights have been proposed in combination with the RNN architecture to improve the forecast performance \cite{lai_modeling_2018,shih_temporal_2019}. 
The sequence-to-sequence (Seq2seq) model has been adopted from natural language processing (NLP) to model sequence data in terms of producing cloud computing load forecasts \cite{peng_multi-step-ahead_2018}.
In addition, WaveNet has been applied for forecasting tasks and has shown good performance in forecasting electric load-demanding tasks \cite{dorado_rueda_short-term_2021}. 
N-beats \cite{oreshkin_n-beats_2020} has demonstrated high performance that outperformed other models in time series competitions, including the M4 and Tourism datasets \cite{makridakis_m4_2020,athanasopoulos_tourism_2011}. Furthermore, the model is designed to include the trend and seasonality decomposition to make it more interpretable.
Additionally, self-attention-based methods like Transformer and Informer have been proposed and demonstrated competitive performance in long-sequence forecasting tasks \cite{li_enhancing_2019,zhou_informer_2021}.

\smallskip
\noindent \textbf{Time series explainability}: 
In eXplainable Artificial Intelligence (XAI), usually, two main categories are considered in terms of model interpretability: inherent model transparency and post-hoc explanations \cite{barredo_arrieta_explainable_2020}. 
In the forecasting setup, classical statistical models like exponential smoothing and ARIMA can be considered inherently transparent, as their parameters can be used to interpret the forecasts using the historical time series patterns \cite{petropoulos_forecasting_2022}. 
However, for complex model architectures, especially for DL forecasting models, it is challenging to use the model parameters to explain the model forecasts directly. In a recent approach to mitigate this challenge, the N-Beats model incorporates multiple trend and seasonality stacks to generate model explainability using a double residual stacking architecture; the constrained trend and seasonality models can decompose the forecasts to human-understandable outputs \cite{oreshkin_n-beats_2020}. 
Additionally, attention mechanisms have been incorporated in LSTM and Transformer models to provide interpretable insights into temporal dynamics \cite{song_attend_2018,lim_temporal_2021}. 
Alternatively, post-hoc explanations have been explored for different forecasting models to provide explanations for the predictions.
Recently, LIMREF was proposed to provide rule explanations for global forecasting models, with a use case of electricity demand prediction \cite{rajapaksha_limref_2022}. Moreover, TS-MULE extended LIME's approach by developing several time series segmentation techniques, including matrix profile and SAX transformation, to identify the important segments from the input time series \cite{schlegel_ts-mule_2021}; FI-SHAP applied the SHAP method for boosting algorithms in providing feature importance and supporting feature engineering for forecasting \cite{zhang_fi-shap_2022}.

\smallskip
\noindent \textbf{Counterfactual explanations for time series}:
Counterfactual explanations were originally proposed to provide sample-based explanations showing which features need to be modified to achieve the desired prediction outcome, and have been adopted in different domain applications like credit risk prediction and breast cancer diagnostic tasks \cite{mothilal_explaining_2020,verma_counterfactual_2020}.  
Specifically, in time series classification (TSC), the concept of counterfactuals was first adopted by Karlsson et al. using a times series tweaking approach to generate counterfactuals for global random shapelet forest models \cite{karlsson_explainable_2018}.
After that, the counterfactual approach was expanded for univariate time series classification using latent space perturbation and instance-based modification \cite{wang_learning_2021,delaney_instance-based_2021}, and multivariate time series classification using heuristic search \cite{ates_counterfactual_2021}. To the best of our knowledge, counterfactual explanations have not been generalized to time series forecasting, especially to multi-horizon forecasting problems. 

\subsection{Problem formulation}

Our objective is to generate time series counterfactuals so that a given black-box forecasting model alters its predictions over a given forecasting time horizon so that these predictions fall within a value range. This value range is defined by a pair of lower and upper bound constraints for each point in the forecasting time horizon.

More formally, let $\boldsymbol{x} = <x_1, x_2, ..., x_n>$ be a \emph{univariate time series} of $n$ timesteps, with each $x_i \in \mathbb{R}$. We also denote as $f(\cdot)$ a black-box time series forecasting model that, given the last $d$ timesteps of $\boldsymbol{x}$, also referred to as \emph{back horizon}, and a forecasting time horizon $T$, it predicts the next $T$ values of the time series, i.e.,
\[
f(<x_{n-d+1}, ..., x_{n-1}, x_n>) = <\hat{x}_{n+1}, \hat{x}_{n+2}, ..., \hat{x}_{n+T}> \ ,
\]
with $\hat{x}_i \in \mathbb{R}$ denoting the predicted value for time point $i$, for $i=n+1, \ldots, n+T$.
Moreover, we define a pair of vectors $\{\boldsymbol{\alpha}=\{\alpha_1, \ldots \alpha_T\}, \boldsymbol{\beta}=\{\beta_1, \ldots \beta_T\}\}$, each of length $T$, that can be used to impose constraints to the forecasted values, which we refer to as \emph{lower bound} and \emph{upper bound} constraints, respectively.  We assume that given $\boldsymbol{x}$ and the forecasting time horizon $T$, the values of these two vectors are defined by two functions, i.e., $\boldsymbol{\alpha}=\gamma_{lb}(\boldsymbol{x}, T)$, $\boldsymbol{\beta}=\gamma_{ub}(\boldsymbol{x}, T)$. These functions can be defined in various ways, and in our experimental evaluation, we demonstrate different instantiations. Hence, the problem studied in this paper is defined as follows.

\smallskip
\noindent
\emph{Problem: }
\textbf{Range-based counterfactuals for time series forecasting. }\label{problem}
Consider a univariate time series sample $\boldsymbol{x} = <x_1, x_2, ..., x_n>$, a back horizon $d$, a forecasting time horizon $T$, the series $\hat{\boldsymbol{x}}=<\hat{x}_{n+1}, \hat{x}_{n+2}, ..., \hat{x}_{n+T}>$ of the next $T$ values predicted by a forecasting model $f$, and a pair of lower and upper bound constraint vectors $\{\boldsymbol{\alpha},\boldsymbol{\beta}\}$. Our goal is to modify $\boldsymbol{x}$ to $\boldsymbol{x}'$ over the back horizon time span $d$, such that the forecasting model produces a new series $\hat{\boldsymbol{x}'} = f(\boldsymbol{x}')$, which we refer to as \emph{counterfactual forecast}, with $\alpha_j \leq \hat{x_i'} \leq \beta_j$, $\forall \hat{x_i'} \in \hat{\boldsymbol{x}'}, i \in [n+1, n+T], j \in [1, T]$, and 
\begin{equation}\label{eq:loss-function}
\boldsymbol{x}' = \underset{\boldsymbol{x}^{*}} {\operatorname{\argmin}} \quad ||f(\boldsymbol{x}^{*}) - \boldsymbol{\alpha}|| + ||\boldsymbol{\beta} - f(\boldsymbol{x}^{*})|| \ .
\end{equation}

In other words, we want to generate a counterfactual $\boldsymbol{x}'$ that ensures that the forecasted values by $f$ are within the defined lower and upper bound constraints for the next $T$ time steps of the forecasting time horizon. Note that the generated counterfactual is considered fully valid when it satisfies these constraints for each of the next $T$ time steps.

\section{ForecastCF: Counterfactual Explanations for Time Series Forecasting}
We propose ForecastCF, an example-based approach to explaining the forecasted values of a black-box forecasting model. Our approach employs gradient-based perturbation for generating the counterfactuals and uses different instantiations of the desired lower and upper bound constraints for the forecasting outcomes.

\subsection{Gradient-based perturbation}
Algorithm~\ref{alg:pseudo-forecastcf} provides the pseudo-code for ForecastCF. Specifically, ForecastCF utilizes gradient descent optimization to perturb the input sample $\boldsymbol{x}$ directly in the input feature space to get the counterfactual $\boldsymbol{x}'$, considering the following loss function:
\begin{equation}
    L = \boldsymbol{v} \odot ( ||f(\boldsymbol{x}^{*})-\boldsymbol{\alpha}|| + ||\boldsymbol{\beta} - f(\boldsymbol{x}^{*})|| )  \ , 
\end{equation}

where $\boldsymbol{v}$ is a \emph{binary masking vector} obtained by a function $\tau(\cdot)$ given the lower and upper bound constraints, i.e.,
\begin{equation}
v_i = \tau(\hat{x}_i^*, \alpha_i, \beta_i) = 
\begin{cases}
0 , if \ \hat{x_i^*} \geq \alpha_i \And \hat{x_i^*} \leq \beta_i,   \\
1 , otherwise. \\ 
\end{cases}       
\end{equation}
for $v_i \in \boldsymbol{v}$, $\alpha_i \in \boldsymbol{\alpha_i}$, and $\beta_i \in \boldsymbol{\beta_i}$ at timestep $i$; and $\hat{x_i^*}$ is the forecast value for the sample $\boldsymbol{x}^{*}$ during the search iteration $it$. The purpose of the binary masking vector is to remove the timesteps that already satisfy the condition in the counterfactual generation for each iteration of the optimization function.

Additionally, we define Adam \cite{kingma_adam_2017} as the optimization function in ForecastCF (see Line~9-13 in Algorithm~\ref{alg:pseudo-forecastcf}). The gradient-based perturbation utilizes the partial derivative of $L$ with respect to the search sample $\boldsymbol{x^{*}}$, as following: 
\begin{equation*}
    \frac{\partial L}{\partial x^{*}} 
        = \boldsymbol{v} \odot (2 * (f(\boldsymbol{x^{*}}) - \boldsymbol{\alpha}) * f'(\boldsymbol{x^{*}}) + 2 * (f(x^{*}) - \boldsymbol{\beta}) * f'(\boldsymbol{x^{*}})) ,
\end{equation*}

where $f'(x^{*})$ is the derivative of the differentiable function $f(\cdot)$ with respect to $\boldsymbol{x^{*}}$. 

Within the while loop defining the constraint conditions in Line~8 (i.e., $\boldsymbol{\hat{x}^{*}} > \boldsymbol{\beta} \lor \boldsymbol{\hat{x}^{*}} < \boldsymbol{\alpha}$ for desired bounds and $t < max\_iter$ for the maximum iteration), we iteratively apply Adam optimization to apply the gradients to the search sample $\boldsymbol{x^{*}}$, where $m_{it}$ and $s_{it}$ are the exponential average of gradients and squares of gradients along $x_{t}^*$ at timestep $t$; in other words, Adam optimizes each timestep $t$ separately with the adaptive learning rate. 
Finally, the output $\boldsymbol{x'}$ is considered a counterfactual of the input sample $\boldsymbol{x}$, when the constraint condition breaks. 
Additionally, the ForecastCF algorithm has two hyperparameters for controlling the counterfactual search convergence: the learning rate $\eta$ and the maximum iteration $max\_iter$; together with three hyperparameters $\gamma_1, \gamma_2, \epsilon$ for the bias corrections of Adam optimization. 

\begin{algorithm}[t]
\caption{ForecastCF search} \label{alg:pseudo-forecastcf}
\SetEndCharOfAlgoLine{}
\SetKwInOut{Input}{input}\SetKwInOut{Output}{output}
\Input{Time series sample $\boldsymbol{x}$, 
differentiable forecast function $f(\cdot)$, 
lower bound $\boldsymbol{\alpha}$, 
upper bound $\boldsymbol{\beta}$, 
learning rate $\eta$, 
maximum iteration $max\_iter$,
Adam hyperparameters $\gamma_1, \gamma_2, \epsilon$
}
\Output{Generated counterfactual $x'$ with desired forecast outcome}
$\boldsymbol{x^{*}} \leftarrow \boldsymbol{x}$ \;

$\boldsymbol{\hat{x}^{*}} \leftarrow f(\boldsymbol{x^{*}})$ \;
$\boldsymbol{v} \leftarrow \tau(\boldsymbol{x^*}, \boldsymbol{\alpha}, \boldsymbol{\beta})$ \;
$loss \leftarrow L(\boldsymbol{x^{*}}, \boldsymbol{\alpha}, \boldsymbol{\beta}, \boldsymbol{v})$ \;
$t \leftarrow 0$ \;
$m_0 \leftarrow 0$ \;
$s_0 \leftarrow 0$ \;

\While { $(\boldsymbol{\hat{x}^{*}}>\boldsymbol{\beta} \lor\boldsymbol{\hat{x}^{*}}<\boldsymbol{\alpha}) \land (t<max\_iter)$}{
    $g_{t} \leftarrow \frac{\partial L}{\partial \boldsymbol{x^{*}}} $ \;
    \For { $j \leftarrow 1 \ to \ T$}{
        $m_{tj} \leftarrow \gamma_1 * m_{tj-1} - (1-\gamma_1) * g_{tj} $  \;
        $s_{tj} \leftarrow \gamma_2 * s_{tj-1} - (1-\gamma_2) * g_{tj}^2 $  \;
        $x_j^{*} \leftarrow x_{j-1}^{*} - \eta\frac{m_{tj}}{\sqrt{s_{tj} + \epsilon}} * g_{tj}$\;        
    }
    $\boldsymbol{\hat{x}^{*}} \leftarrow f(\boldsymbol{x^{*}})$ \;
    $\boldsymbol{v} \leftarrow \tau(\boldsymbol{x^*}, \boldsymbol{\alpha}, \boldsymbol{\beta})$ \;
    $loss \leftarrow L(\boldsymbol{x^{*}}, \boldsymbol{\alpha}, \boldsymbol{\beta}, \boldsymbol{v})$ \;
    $t \leftarrow t + 1$ \;
}

$\boldsymbol{x'} \leftarrow \boldsymbol{x^{*}}$ \;
\Return{$\boldsymbol{x'}$}
\vspace{-1mm}
\end{algorithm}

\subsection{Instantiations of desired trajectory bounds}
In this section, we formulate two instantiations for desired trajectory bounds in our proposed algorithm. We show that the desired trajectory can be customized by a range of hyperparameters to provide the upper bound vector $\boldsymbol{\alpha}$ and the lower bound vector $\boldsymbol{\beta}$. Note that these hyperparameters can be adjusted for particular scenarios. 

\smallskip
\noindent \textbf{Polynomial trend.} 
In the polynomial trend instantiation, we define five hyperparameters to choose the desired prediction outcome:
center function $c(\cdot)$, 
shift $s$, fraction of standard deviation $fr$, 
the desired change percent $cp$, and the polynomial order $poly\_order$. 

Center function $c(\cdot)$ and shift $s$ are defined to choose the starting value of the ``trajectory bounds''; the fraction $fr$ is to define the width of the bounds; more specifically, for an input sample $\boldsymbol{x}$, the starting values are calculated as 
\begin{equation*}
\begin{split}
    \alpha_1 = c(x) * (1 + s - fr * \sigma(x)) , \\
    \beta_1 = c(x) * (1 + s + fr * \sigma(x)) , 
\end{split}
\vspace{-2mm}
\end{equation*}
for the upper and lower bounds separately, where $c(\cdot)$ can be defined as either median, max, min, mean, or the last point of $\boldsymbol{x}$, $\sigma(x)$ is the standard deviation of the input sample.   
While the ending values of the trajectory bounds $\alpha_T$ and $\beta_T$ can be defined as:
\begin{equation*}
\begin{split}
    \alpha_T = \alpha_1 +  c(x) * (1 + s + cp) , \\
    \beta_T = \beta_1 +  c(x) * (1 + s + cp) .
\end{split}
\vspace{-2mm}
\end{equation*}

With the starting and ending values of the desired bounds, we fit a polynomial function of order $poly\_order$, over the forecasting time horizon $T$ and hence obtain the lower and upper bound vectors $\boldsymbol{\alpha}, \boldsymbol{\beta}$. 
Note that when $cp$=$0$, the trajectory bounds become two horizontal bounds, which can be suitable for a task of stabilising the forecasting values in the desired range.

\smallskip
\noindent \textbf{Polynomial trend with limitations.}
The second instantiation we propose is a variant of the polynomial trend instantiation. In addition to the hyperparameters in the polynomial trend, we define two hyperparameter constraints $l_{\alpha}$ and $l_{\beta}$ that replace the bound values that exceed these constraint values for both $\boldsymbol{\alpha}$
and $\boldsymbol{\beta}$. More specifically, 
\begin{equation*}
\alpha_t = 
\begin{cases}
    l_{\alpha},\  if \ \alpha_t \leq l_{\alpha}, \\
    \alpha_t, otherwise,
\end{cases}
and \  
\beta_t = 
\begin{cases}
l_{\beta}, \ if \ \beta_t \geq l_{\beta}, \\
\beta_t, otherwise,
\end{cases}
\end{equation*}

for both $\alpha_t \in \boldsymbol{\alpha}$ and $\beta_t \in \boldsymbol{\beta}$ at timestep $t$ in the defined trajectory bounds. This instantiation can be applied to medical prognostic tasks where the patient's condition is required to be stabilised after a desired shift, e.g., hypotension prediction.

As a side note, the shape of constraint bounds is determined by using a self-defined function in the ForecastCF algorithm; the previous two instantiations show that this function can be defined as the polynomial function, and polynomial with additional constraints. Without loss of generality, the self-defined function can be modified to other forms of the desired trajectory range, incorporating trend and seasonality information in the desired counterfactual outcome, e.g., exponential or trigonometric functions.

\section{Experimental Evaluation}
\subsection{Data preparation}
\noindent \textbf{Data sources.} 
We evaluate our proposed ForecastCF algorithm on four benchmark datasets from recent forecasting competitions: \emph{CIF2016} \cite{stepnicka_results_2017}, \emph{NN5} \cite{ben_taieb_review_2012}, \emph{Tourism} \cite{athanasopoulos_tourism_2011}, and \emph{M4~Finance} \cite{makridakis_m4_2020}; together with two additional datasets in stock marketing and healthcare: \emph{SP500}, and \emph{MIMIC} \cite{johnson_mimic-iii_2016}.
All competition datasets contain univariate time series from different sources:
\emph{CIF2016} consists of real-world banking data, and the forecast is to forecast 12 months ahead from 57 series with different lengths;
\emph{NN5} is a 56-day-ahead forecast task with 111 daily ATM cash withdrawal data;
\emph{Tourism} includes monthly series with different lengths from 366 countries;
\emph{M4~Finance} is extracted from the M4 competition \cite{makridakis_m4_2020} with a focus on the monthly financial forecasting task, containing 10,987 series with different lengths.  
Additionally, we introduce the \emph{SP500} dataset, originally collected from Yahoo Finance\footnote[4]{https://finance.yahoo.com/}, including the Open stock prices of companies based on the S\&P500 index between 2013-02-08 and 2018-02-07. 
For the \emph{MIMIC} dataset, we extract the mean arterial pressure (MAP) measurements from 1,035 cardiovascular patients from the first 48 hours since their admission in the MIMIC-III database \cite{johnson_mimic-iii_2016}. 
We adopt the task of hypotension prediction \cite{hatib_machine-learning_2018} to forecast an 8-hour-ahead MAP based on the historical 24-hour measurement. 
The summary statistics are described in Table~\ref{tab:data-stastics}. 

\begin{table}[tb]\centering
\caption{Statistics of the datasets}\label{tab:data-stastics}
\vspace{-2mm}
\begin{tabular}{l |r|r|r|r|r}
\hline
Dataset          &\#samples  &max\_{len} &min\_{len} &mean\_{len} &horizon \\ \hline
\textbf{CIF2016}    &57     &120    &60     &112     &12 \\
\textbf{NN5}        &111    &791    &791    &791     &56 \\
\textbf{Tourism}    &366    &333    &91     &299     &24 \\
\textbf{M4 Finance} &10,987 &1502   &60     &184     &18 \\ 
\textbf{SP500}      &468    &1259   &1259   &1259    &60 \\ 
\textbf{MIMIC}      &1,035  &48     &48     &48      &8 \\ \hline
\end{tabular}
\vspace{-2mm}
\end{table}

\begin{figure}[tb]
\centerline{\includegraphics[width=0.65\linewidth]{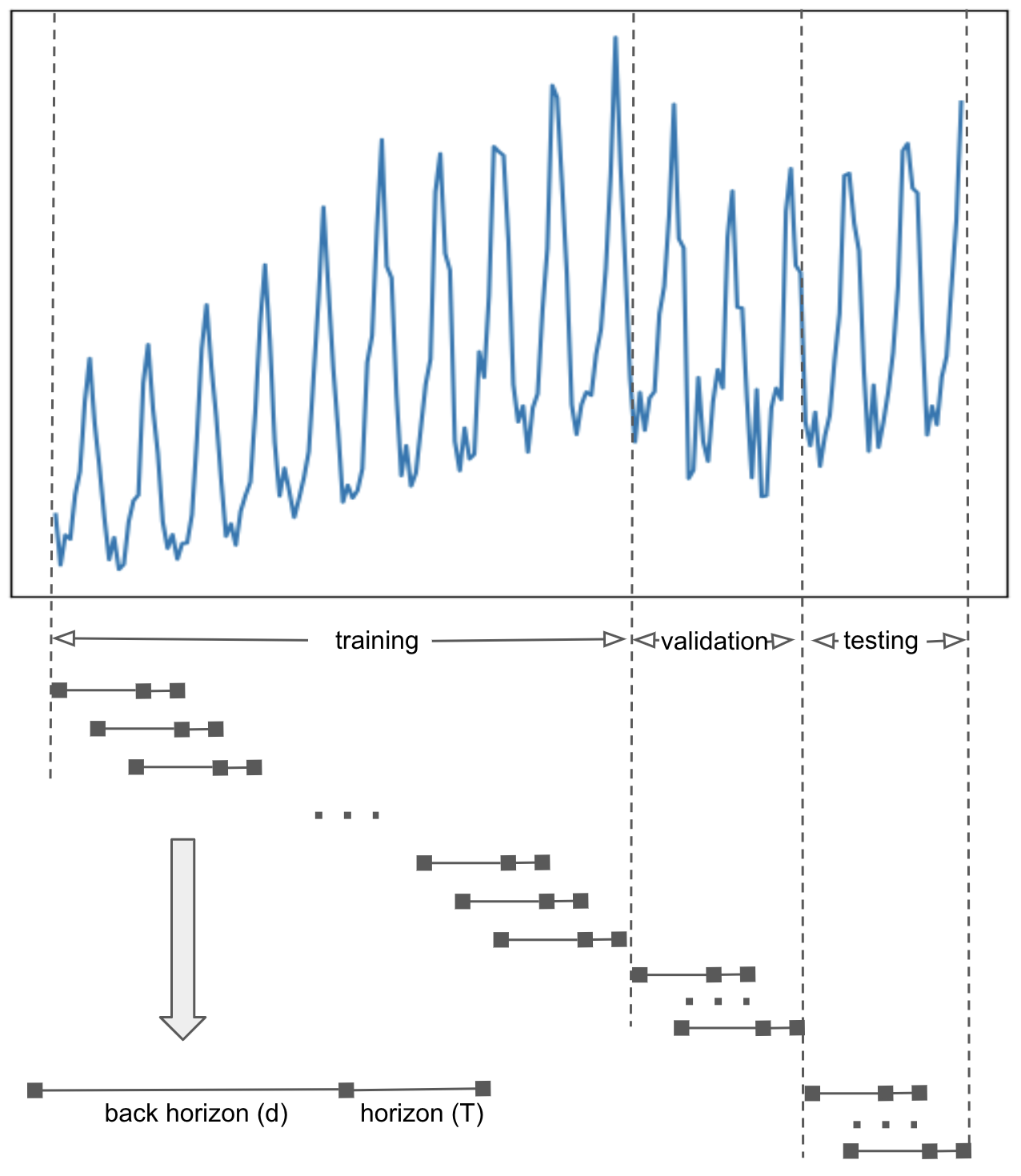}}
\caption{The data-preprocessing setup: each individual time series is split into training, validation and testing chunks, and then divided into smaller sequences, consisting of `back horizon' and `horizon' for evaluation purposes.}
\label{fig:train-test-split}
\vspace{-4mm}
\end{figure}

\smallskip
\noindent \textbf{Data-preprocessing.} 
We follow the constant in-sample strategy for the rolling origin evaluation technique \cite{tashman_out--sample_2000}, as shown in Fig.~\ref{fig:train-test-split}. 
Specifically, we utilize the temporal order to split each individual time series into training, validation, and testing chunks, based on the split size (e.g., 60\%/20\%/20\%), and then we aggregate the individual chunks into the final training, validation, and test sets for each dataset. For notation purposes, we use $\boldsymbol{\hat{\mathcal{X}}}$ to denote the test set and $\boldsymbol{\hat{\mathcal{X}'}}$ to denote the counterfactuals generated for each example in the test set. Moreover, each chunk is further divided into two subsequences (i.e., `back horizon' and `forecasting time horizon') using a stride step of the rolling origin, for the purpose of training DL models and generating counterfactual explanations. 
Except for \emph{MIMIC}, we directly apply the rolling origin approach with a stride step $8$ to each time series (i.e., the horizon of each patient) to split into subsequences before aggregating them. Among all the datasets, a standard min-max normalization is applied using scales of the training set.

\subsection{Experimental setup}
We evaluate ForecastCF on four DL model architectures, including state-of-the-art models benchmarked for univariate time series forecasting: 
\begin{itemize}
    \item \emph{GRU}: a GRU model consists of two consecutive GRU layers with 100 hidden units separately.
    \item \emph{Seq2seq}\cite{peng_multi-step-ahead_2018}: based on the encoder-decoder framework in NLP, we adopt the TFTS implementation\footnote[1]{\label{note1}https://github.com/LongxingTan/Time-series-prediction} by utilizing 256 units in RNN layers and 256 units for the dense layer. 
    \item \emph{WaveNet}\cite{dorado_rueda_short-term_2021}: adoption of the WaveNet model, which can capture autocorrelations and correlations with related time series with dilated convolutional layers.
    We adopt TFTS\ref{note1} and then modify the default architecture to include the skip connection with the filter size 256.
        \item \emph{N-Beats}\cite{oreshkin_n-beats_2020}: consists of double residual stacks of fully connected layers, considered as current state-of-the-art in forecasting benchmarks.
    We apply the N-Beats implementation\footnote[7]{https://github.com/philipperemy/n-beats}, where we set the dense layer to 256 hidden units, and keep the remaining default hyperparameters. 
\end{itemize}

In the experiment, we apply a constant factor of the forecast horizon to determine the back horizon for training the DL models (i.e., back\_horizon = constant * horizon).
Specifically, for \emph{CIF2016}, \emph{NN5}, and \emph{Tourism} we apply a constant of $1.25$ following earlier studies \cite{bandara_forecasting_2020,hewamalage_recurrent_2021}; for \emph{M4~Finance} we set the constant to $1.5$ \cite{oreshkin_n-beats_2020}; and for \emph{SP500} and \emph{MIMIC} the constant is set to $2$ and $3$, respectively. 
Finally, each forecasting model is trained with the loss of mean absolute error (MAE) using an Adam optimizer (with learning rate 0.0001). The batch size is set to 128 and the number of training epochs is 100; early stopping is applied to the validation loss to prevent over-fitting. 

Next, we evaluate the performance of the counterfactual explanations generated by the ForecastCF algorithm compared to two baseline approaches.
In ForecastCF, we choose the learning rate $\eta$=$0.001$ and $max\_iter$=$100$ after an empirical hyperparameter search, and Adam hyperparameters $\gamma_1, \gamma_2, \epsilon$ are set to default in the keras implementation.

\smallskip
\noindent \textbf{Baselines.}
We additionally compare ForecastCF with two baseline models for the counterfactual generation: 
\begin{itemize}
    \item \emph{BaseNN}: it applies the 1-nearest-neighbour approach to retrieve the nearest neighbour from the training set as the counterfactual. It utilizes the Euclidean distance between the desired forecast range (average between the upper and lower bounds) and the values of the forecasting time horizon from the training set. 
    \item \emph{BaseShift}: it utilizes a naive shifting technique by directly multiplying the shift factor (i.e., the desired percentage change) with the input series to get the counterfactual.
\end{itemize}

Finally, we conduct three experiments to investigate the effectiveness of our proposed ForecastCF: 
(1) applying the \emph{polynomial trend} instantiation ($center$=$median$, shift $s$=$0$ and fraction $fr$ determined from an empirical search for each dataset) for four forecasting models using the defined horizon of each dataset (see Table~\ref{tab:data-stastics});
(2) investigating counterfactual performance when modifying the forecast horizon of four DL models gradually, ranging from $1$ to the defined horizon; and 
(3) an ablation study on two hyperparameters from ForecastCF on \emph{CIF2016}: desired change percentage $cp$ (between -$50\%$ and +$50\%$) and fraction $fr$ (from $0.25$ to $5$). 
All the experiments and the computational runtime are evaluated with NVIDIA GeForce RTX 2080 (GPU) and AMD Ryzen Threadripper 2950X 16-Core Processor (CPU).

\subsection{Evaluation Metrics}
\smallskip
\noindent \textbf{Predictive performance. }
For evaluating the predictive performance of the forecasting models, we apply two commonly used measures \cite{makridakis_m4_2020,athanasopoulos_tourism_2011}, Symmetric Mean Absolute Percentage Error (\emph{sMAPE}) and Mean Absolute Scaled Error (\emph{MASE}). 

\emph{sMAPE} measures the scaled error between the forecast and the ground truth:
\begin{equation*}
    \emph{sMAPE} = 200 * \frac{1}{T}\sum_{i=1}^{T}
    \frac{|x_{d+i} - \hat{x}_{d+i}|}{|x_{d+i}| + |\hat{x}_{d+i}|}.
\end{equation*}

\emph{MASE} measures the scaled error between the forecast and the forecast obtained with a naive forecast model:
\begin{equation*}
    \emph{MASE} = \frac{1}{T}\sum_{i=1}^{T}
    \frac{|x_{d+i} - \hat{x}_{d+i}|}{
    \frac{1}{d+T-m}\sum_{j=m+1}^{d+T}|x_j - x_{j-m}|}, 
\end{equation*}
where $T$ is the forecasting time horizon, $d$ is the back horizon, and $m$ is the periodicity of the data (e.g., $m$=$1$ for the naive forecaster directly using the previous step). For sMAPE, a lower score is better; and MASE $<1$ indicates that the forecast performs better than the naive forecaster, and vice versa.
We report the average of all test samples for each dataset over five repetition runs.

\smallskip
\noindent \textbf{Counterfactual evaluation. }
We adopt evaluation metrics from recent counterfactual studies to the forecasting setup. We consider two groups of metrics: (a) \emph{validity} and (b) \emph{data manifold closeness}. 

\noindent
(a) We measure \emph{validity} through two metrics, \emph{Validity Ratio} and \emph{Stepwise Validity AUC}.

\emph{Validity Ratio}: we follow the validity metric defined in earlier works \cite{verma_counterfactual_2020,mothilal_explaining_2020}, and reformulate it into a ratio score that computes the average over all counterfactuals of the proportion of valid timesteps of the counterfactual forecasts, i.e.,
\begin{equation*} \label{eq1}
\emph{Ratio}(\boldsymbol{\hat{\mathcal{X}}'}) = \frac{1}{K}\sum_{k=1}^{K} 
(\frac{1}{T}\sum_{i=1}^{T} \tau_b(\hat{x_i'}^k, \alpha^k_{i}, \beta^k_{i})),  
\forall \boldsymbol{\hat{x'}}^k \in \boldsymbol{\hat{\mathcal{X}'}},
\end{equation*}
where $K$ is the number of generated counterfactuals in $\boldsymbol{\hat{\mathcal{X}'}}$, and $\tau_b$ is a function that measures the validity of the $i$-th forecasted value $\hat{x_i'}$ of a given counterfactual $\boldsymbol{x'}$, i.e.,
\begin{equation*}
\tau_b(\hat{x_i'}, \alpha_i, \beta_i) = 
\begin{cases}
1 , if \ \alpha_i \leq  \hat{x_i'} \leq \beta_i ,   \\
0 , otherwise. \\ 
\end{cases}       
\end{equation*}

\emph{Stepwise Validity AUC}: we additionally propose a novel metric to measure the area under the curve (AUC) of function $\phi(\cdot)$ that computes the fraction of counterfactuals with $t$ consecutively valid forecasted values (y-axis) over the corresponding fraction $t/T$ of the forecasting time horizon $T$ (x-axis).   
More formally, function $\phi(\cdot)$ is defined as follows: 
\begin{equation*} \label{eq3}
 \phi(t) = \frac{1}{K}|\{\boldsymbol{\hat{x'}}^k \in \boldsymbol{\hat{\mathcal{X}'}}:  
 \sum_{i=1}^{T} \tau_c(\hat{x_i'}^k, \alpha^k_{i}, \beta^k_{i})) = t\}|,
\end{equation*}
where $\tau_c$ is a function that measures the validity of the $i$-th forecasted value $\hat{x_i'}$ of a given counterfactual $\boldsymbol{x'}$, i.e.,
\begin{equation*}
\tau_c(\hat{x_i'}, \alpha_i, \beta_i) = 
\begin{cases}
1 , if \ \alpha_i \leq  \hat{x_i'} \leq \beta_i \And \tau_c(\hat{x'}_{i-1})=1,   \\
0 , otherwise. \\ 
\end{cases}       
\end{equation*}
Note for $i=1$ we apply $\tau_b(\cdot)$ instead of $\tau_c(\cdot)$ for counting the first valid timestep.

Thus, Stepwise Validity AUC is defined as follows:
\begin{equation*} \label{eq2}
\emph{Step-AUC}(\boldsymbol{\hat{\mathcal{X}}'}) = 
\int_{t=0}^{1} \phi(t) dt .
\end{equation*}
For both \emph{Validity Ratio} and \emph{Stepwise Validity AUC}, a higher score (i.e., closer to $1$) indicates a higher fraction of valid counterfactual forecast points, hence better performance.

\noindent
(b) We measure \emph{data manifold closeness} through two metrics, \emph{Proximity} and \emph{Compactness}.

\emph{Proximity}: 
the proximity metric indicates the average Euclidean distance between the original samples and the counterfactual samples \cite{delaney_instance-based_2021,wang_learning_2021}, where lower proximity is desired. More formally,
\begin{equation*} \label{eq:proximity}
\emph{Proximity}(\boldsymbol{\mathcal{X}'}) = 
\frac{1}{K}\sum_{k=1}^K 
||\boldsymbol{x^k} - \boldsymbol{x'}^k|| , \forall \boldsymbol{x'}^k \in \boldsymbol{\mathcal{X}'}.
\end{equation*}

\emph{Compactness}: compactness (also known as `sparsity') measures the average proportion of timesteps that remain similar to the original samples \cite{delaney_instance-based_2021,karlsson_explainable_2018}, where higher compactness is desired. More formally,
\begin{equation*} \label{eq4}
\emph{Compact}(\boldsymbol{\mathcal{X}'}) = \frac{1}{K}\sum_{k=1}^{K}(
\frac{1}{T}\sum_{i=1}^T
c ({x_{i}'}^k, tol) ), 
\forall \boldsymbol{x'}^k \in \boldsymbol{\mathcal{X}'},
\end{equation*}
where 
\begin{equation*}
c (x_i', tol) = 
\begin{cases}
1 , |x_i - x_i'| \leq tol,   \\
0 , otherwise, \\ 
\end{cases}       
\end{equation*}
with $tol$ being the tolerance between $x_i'$ and the corresponding sample $x_i$ at timestep $i$.

\begin{table*}[t]
\caption{Forecasting model performance for six real-world datasets. We report average scores over five repetition runs.}\label{tab:forecast-metrics}
\begin{center}
\vspace{-4mm}
\begin{tabular}{c|c c|c c|c c|c c|c c|c c}
\hline
 &\multicolumn{2}{|c|}{\textbf{CIF2016}} &\multicolumn{2}{|c|}{\textbf{NN5}} &\multicolumn{2}{|c|}{\textbf{Tourism}} &\multicolumn{2}{|c|}{\textbf{M4 Finance}} &\multicolumn{2}{|c|}{\textbf{SP500}} &\multicolumn{2}{|c}{\textbf{MIMIC}} \\
\textbf{Model} & \textit{sMAPE}& \textit{MASE} & \textit{sMAPE}& \textit{MASE} & \textit{sMAPE}& \textit{MASE} & \textit{sMAPE}& \textit{MASE} & \textit{sMAPE}& \textit{MASE} & \textit{sMAPE}& \textit{MASE} \\
\hline
GRU 
& 11.150 & 1.544 & 29.694 & 0.719 & 35.228 & 1.208 & 12.449 & 7.021 & 11.552 & 13.688 & 9.430 & 1.138\\
Seq2seq 
& 11.457 & 1.625 & 37.458 & 0.922 & 37.288 & 1.314 & 12.219 & 6.592 & 12.276 & 14.453 & \textbf{9.416} & \textbf{1.135}\\
WaveNet 
& \textbf{9.056} & \textbf{1.315} & 35.364 & 0.837 & 25.610 & 0.895 & 13.839 & 6.544 & 8.298 & 9.410 & 10.963 & 1.307\\
N-Beats 
& 10.968 & 1.637 & \textbf{29.385} & \textbf{0.717} & \textbf{25.050} & \textbf{0.843} & \textbf{10.960} & \textbf{5.354} & \textbf{7.477} & \textbf{8.628} & 9.510 & 1.147\\
\hline
\end{tabular}
\vspace{-4mm}
\end{center}
\end{table*}

\begin{table*}[tb]
\caption{Validity: \emph{validity ratio} and \emph{stepwise validity AUC}. We report the average of five runs and highlight the best metric in bold.}\label{tab:validity}
\begin{center}
\vspace{-5mm}
\begin{tabular}{c c|c c|c c|c c|c c|c c|c c}
\hline
& &\multicolumn{2}{|c|}{\textbf{CIF2016}} &\multicolumn{2}{|c|}{\textbf{NN5}} &\multicolumn{2}{|c|}{\textbf{Tourism}} &\multicolumn{2}{|c|}{\textbf{M4 Finance}} &\multicolumn{2}{|c|}{\textbf{SP500}} &\multicolumn{2}{|c}{\textbf{MIMIC}} \\
\textbf{Model} & \textbf{CF model} & \textit{Ratio}& \textit{S-AUC} & \textit{Ratio}& \textit{S-AUC} & \textit{Ratio}& \textit{S-AUC} & \textit{Ratio}& \textit{S-AUC} & \textit{Ratio}& \textit{S-AUC} & \textit{Ratio}& \textit{S-AUC} \\
\hline
GRU & BaseNN 
& 0.474 & 0.244 & 0.879 & 0.477 & 0.910 & 0.751 & 0.601 & 0.557 & 0.276 & 0.190 & 0.639 & 0.505\\
 & BaseShift 
& 0.506 & 0.360 & 0.699 & 0.163 & 0.725 & 0.380 & 0.576 & 0.367 & 0.339 & 0.090 & 0.550 & 0.327\\
 & ForecastCF 
& \textbf{0.781} & \textbf{0.650} & \textbf{1.000} & \textbf{0.980} & \textbf{0.990} & \textbf{0.922} & \textbf{0.832} & \textbf{0.655} & \textbf{0.688} & \textbf{0.338} & \textbf{0.941} & \textbf{0.800}\\\hline
Seq2seq & BaseNN 
& 0.536 & 0.308 & 0.974 & 0.942 & 0.932 & 0.836 & 0.634 & 0.588 & 0.279 & 0.220 & 0.671 & 0.549\\
 & BaseShift 
& 0.487 & 0.318 & 0.939 & 0.842 & 0.790 & 0.459 & 0.579 & 0.419 & 0.270 & 0.076 & 0.502 & 0.327\\
 & ForecastCF 
& \textbf{0.792} & \textbf{0.667} & \textbf{0.995} & \textbf{0.973} & \textbf{0.998} & \textbf{0.953} & \textbf{0.833} & \textbf{0.686} & \textbf{0.557} & \textbf{0.320} & \textbf{0.912} & \textbf{0.760}\\\hline
WaveNet & BaseNN 
& 0.216 & 0.008 & 0.721 & 0.101 & 0.869 & 0.600 & 0.651 & 0.529 & 0.277 & 0.056 & 0.483 & 0.087\\
 & BaseShift 
& 0.332 & 0.043 & 0.552 & 0.024 & 0.622 & 0.117 & 0.530 & 0.149 & 0.224 & 0.026 & 0.382 & 0.043\\
 & ForecastCF 
& \textbf{0.742} & \textbf{0.636} & \textbf{0.997} & \textbf{0.916} & \textbf{0.958} & \textbf{0.691} & \textbf{0.867} & \textbf{0.781} & \textbf{0.933} & \textbf{0.857} & \textbf{0.887} & \textbf{0.713}\\\hline
N-Beats & BaseNN 
& 0.531 & 0.291 & 0.885 & 0.375 & 0.924 & 0.705 & 0.650 & 0.552 & 0.325 & 0.149 & 0.634 & 0.398\\
 & BaseShift 
& 0.362 & 0.055 & 0.655 & 0.052 & 0.630 & 0.174 & 0.482 & 0.240 & 0.296 & 0.039 & 0.532 & 0.221\\
 & ForecastCF 
& \textbf{0.699} & \textbf{0.567} & \textbf{1.000} & \textbf{0.980} & \textbf{0.984} & \textbf{0.920} & \textbf{0.884} & \textbf{0.778} & \textbf{0.879} & \textbf{0.727} & \textbf{0.928} & \textbf{0.772}\\\hline
\end{tabular}
\vspace{-4mm}
\end{center}
\end{table*}

\begin{table*}[tb]
\caption{Data manifold closeness: \emph{proximity} and \emph{compactness}. We report the average of five runs and highlight the best metric in bold. The \protect\footnote[2]{} sign indicates that BaseNN and BaseShift generated the same counterfactuals across different forecasting models due to the nature of the methods. }\label{tab:closeness}
\begin{center}
\vspace{-4mm}
\begin{tabular}{c c|c c|c c|c c|c c|c c|c c}
\hline
& &\multicolumn{2}{|c|}{\textbf{CIF2016}} &\multicolumn{2}{|c|}{\textbf{NN5}} &\multicolumn{2}{|c|}{\textbf{Tourism}} &\multicolumn{2}{|c|}{\textbf{M4 Finance}} &\multicolumn{2}{|c|}{\textbf{SP500}} &\multicolumn{2}{|c}{\textbf{MIMIC}} \\
\textbf{Model} & \textbf{CF model} & \textit{Proxi.}& \textit{Compa.} & \textit{Proxi.}& \textit{Compa.} & \textit{Proxi.}& \textit{Compa.} & \textit{Proxi.}& \textit{Compa.} & \textit{Proxi.}& \textit{Compa.} & \textit{Proxi.}& \textit{Compa.} \\
\hline
GRU & BaseNN\footnote[2] 
& 1.518 & 0.018 & 2.001 & 0.037 & 2.387 & 0.032 & 1.265 & 0.064 & 3.448 & 0.033 & 1.732 & 0.026\\
 & BaseShift\footnote[2]
& 0.265 & 0.091 & \textbf{0.384} & 0.039 & 0.472 & 0.025 & 0.424 & 0.070 & 0.999 & 0.043 & 0.264 & 0.090\\
 & ForecastCF 
& \textbf{0.171} & \textbf{0.534} & 0.503 & \textbf{0.348} & \textbf{0.323} & \textbf{0.514} & \textbf{0.172} & \textbf{0.837} & \textbf{0.660} & \textbf{0.900} & \textbf{0.153} & \textbf{0.846}\\\hline
Seq2seq & BaseNN\footnote[2] 
& 1.518 & 0.018 & 2.001 & 0.037 & 2.387 & 0.032 & 1.265 & 0.064 & 3.448 & 0.033 & 1.732 & 0.026\\
 & BaseShift\footnote[2] 
& 0.265 & 0.091 & 0.384 & 0.039 & 0.472 & 0.025 & 0.424 & 0.070 & 0.999 & 0.043 & 0.264 & 0.090\\
 & ForecastCF 
& \textbf{0.136} & \textbf{0.593} & \textbf{0.012} & \textbf{0.979} & \textbf{0.148} & \textbf{0.653} & \textbf{0.144} & \textbf{0.844} & \textbf{0.704} & \textbf{0.875} & \textbf{0.163} & \textbf{0.839}\\\hline
WaveNet & BaseNN\footnote[2] 
& 1.518 & 0.018 & 2.001 & 0.037 & 2.387 & 0.032 & 1.265 & 0.064 & 3.448 & 0.033 & 1.732 & 0.026\\
 & BaseShift\footnote[2] 
& \textbf{0.265} & 0.091 & \textbf{0.384} & 0.039 & \textbf{0.472} & 0.025 & 0.424 & 0.070 & 0.999 & 0.043 & \textbf{0.264} & 0.090\\
 & ForecastCF 
& 0.340 & \textbf{0.613} & 0.912 & \textbf{0.645} & 0.635 & \textbf{0.705} & \textbf{0.141} & \textbf{0.767} & \textbf{0.421} & \textbf{0.758} & 0.408 & \textbf{0.723}\\\hline
N-Beats & BaseNN\footnote[2] 
& 1.518 & 0.018 & 2.001 & 0.037 & 2.387 & 0.032 & 1.265 & 0.064 & 3.448 & 0.033 & 1.732 & 0.026\\
 & BaseShift\footnote[2] 
& \textbf{0.265} & 0.091 & \textbf{0.384} & 0.039 & \textbf{0.472} & 0.025 & 0.424 & 0.070 & 0.999 & 0.043 & 0.264 & 0.090\\
 & ForecastCF 
& 0.306 & \textbf{0.337} & 0.655 & \textbf{0.080} & 0.562 & \textbf{0.162} & \textbf{0.131} & \textbf{0.589} & \textbf{0.512} & \textbf{0.299} & \textbf{0.183} & \textbf{0.615}\\\hline
\end{tabular}
\vspace{-4mm}
\end{center}
\end{table*}

\subsection{Results}

Table~\ref{tab:forecast-metrics} shows the performance metrics of the four forecasting models in need of explanations. First, we observed that N-Beats obtained the optimal sMAPE and MASE scores among four datasets (NN5, Toursim, M4 and SP500), demonstrating competitive performance compared to the other three forecasting models. In comparison, the WaveNet model and the Seq2seq model outperformed the others in the CIF2016 and MIMIC datasets, respectively. 
We additionally observed that early stopping was activated during the training process, indicating that all the forecasting models converged.

In the first experiment, we applied the \emph{polynomial trend} instantiation to investigate the general performance of counterfactuals by ForecastCF, in comparison with the baseline models. 
In Table~\ref{tab:validity} and \ref{tab:closeness}, we report the average evaluation metrics on the three counterfactual methods for all forecasting models across six different datasets, over five random repetition runs. 
In terms of \emph{validity}, we observed that ForecastCF achieved the highest validity ratio and stepwise validity AUC scores (mostly above 75\%) in all comparisons, suggesting that ForecastCF could generate counterfactuals satisfying a high proportion of desired forecast outcomes consecutively. While the baselines obtained lower validity scores, especially for the SP500 dataset, they had lower than 35\% in both validity ratio and stepwise validity AUC. 
For \emph{data manifold closeness}, we found that ForecastCF obtained the highest compactness for all the datasets, indicating that it could generate more compact counterfactuals compared to the baselines. For proximity, ForecastCF outperformed the other baselines in M4, SP500 and MIMIC datasets; while ForecastCF and BaseShift had similar winning counts for the other three datasets. This evidence suggests that ForecastCF could provide more proximate counterfactuals corresponding to the original samples; while BaseShift could generate relatively close counterfactual samples and BaseNN failed to get proximate counterfactuals. 

\begin{figure}[t] 
\centering
\subfloat[CIF2016\label{subfig:1a}]{%
   \includegraphics[width=0.48\linewidth]{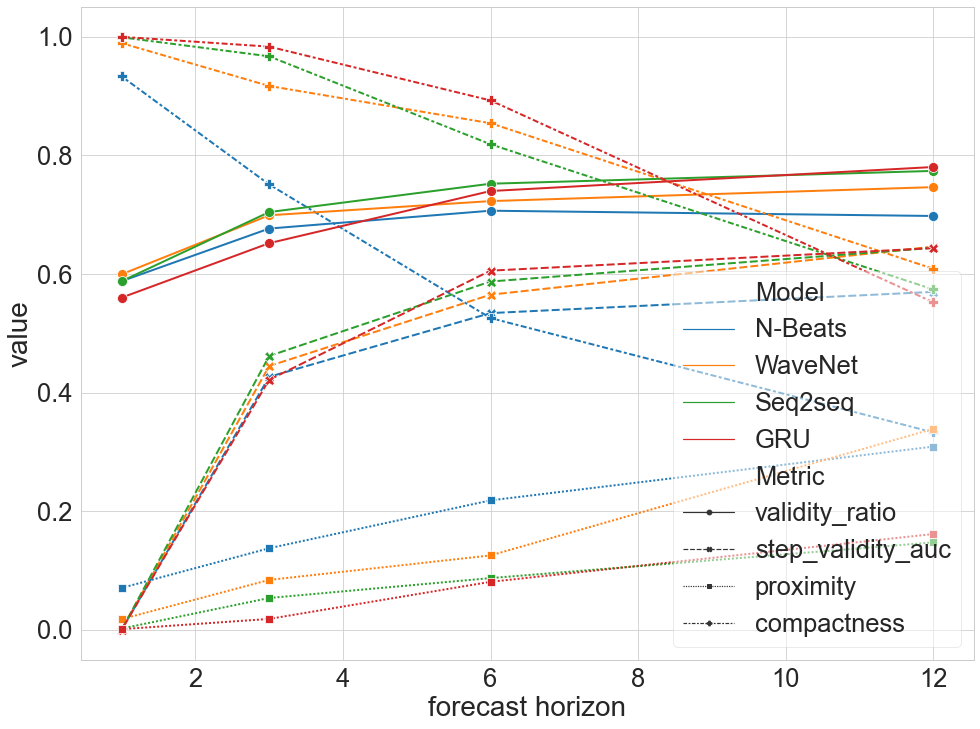}}
   \vspace{-2mm}
\hfill
\subfloat[NN5\label{subfig:1b}]{%
    \includegraphics[width=0.48\linewidth]{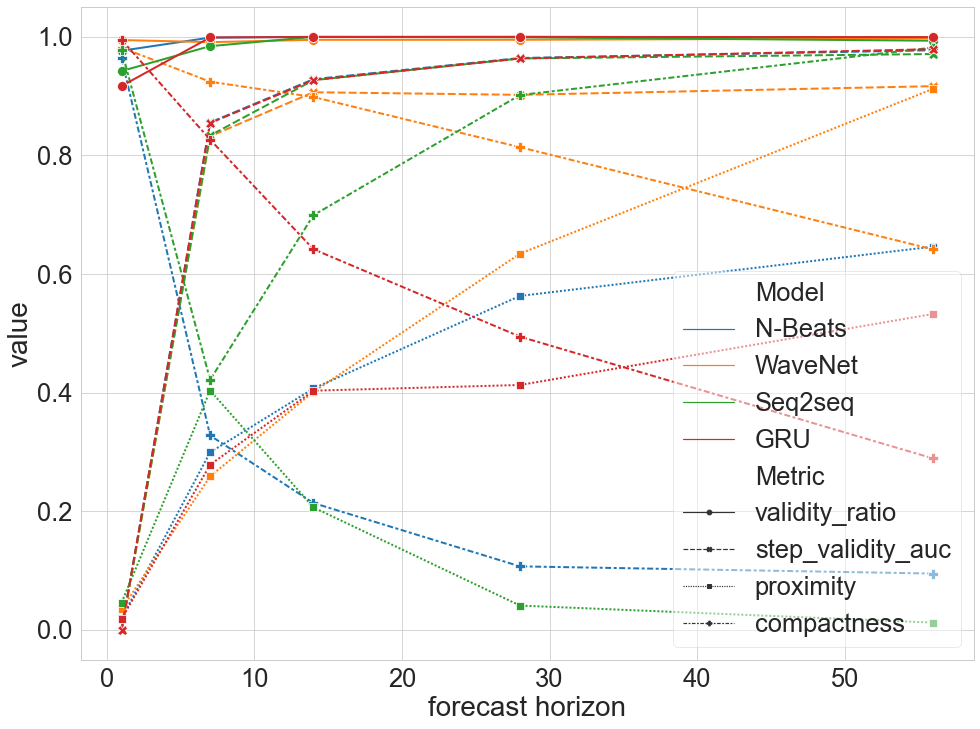}}
\\
\subfloat[Tourism\label{subfig:1c}]{%
   \includegraphics[width=0.48\linewidth]{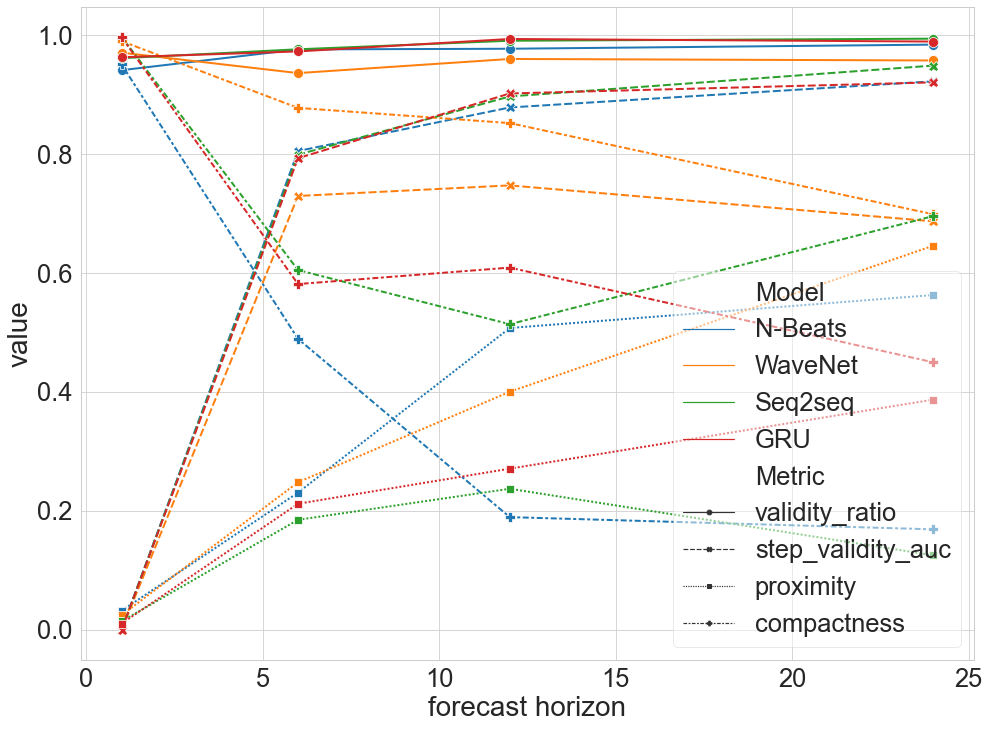}}
   \vspace{-2mm}
\hfill
\subfloat[M4 Finance\label{subfig:1d}]{%
    \includegraphics[width=0.48\linewidth]{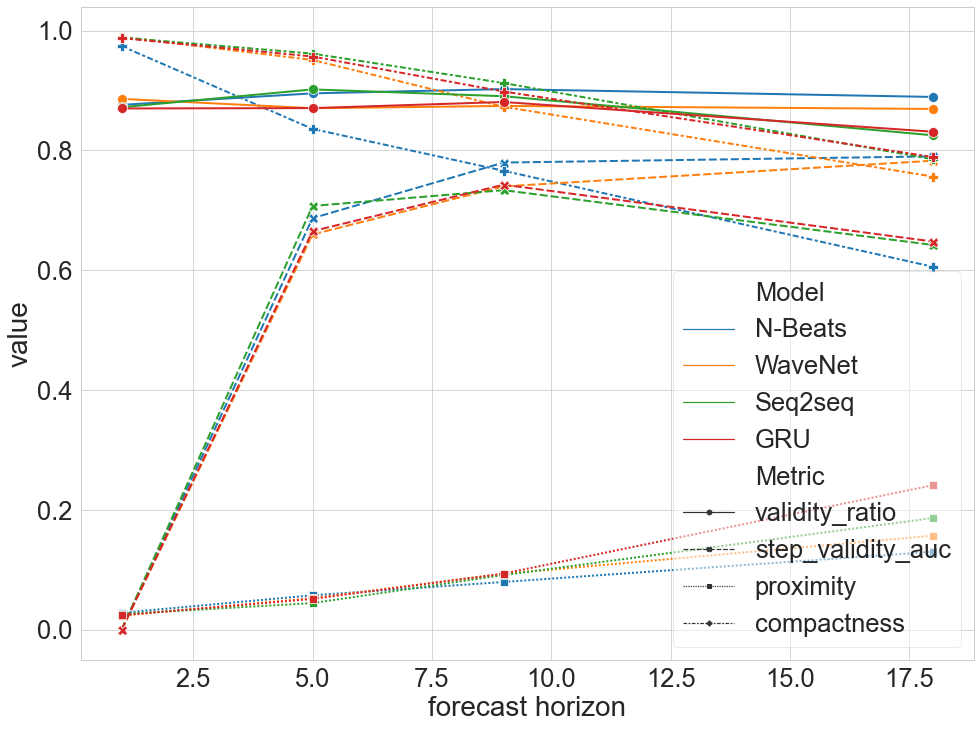}}
\\
\subfloat[SP500\label{subfig:1e}]{%
    \includegraphics[width=0.48\linewidth]{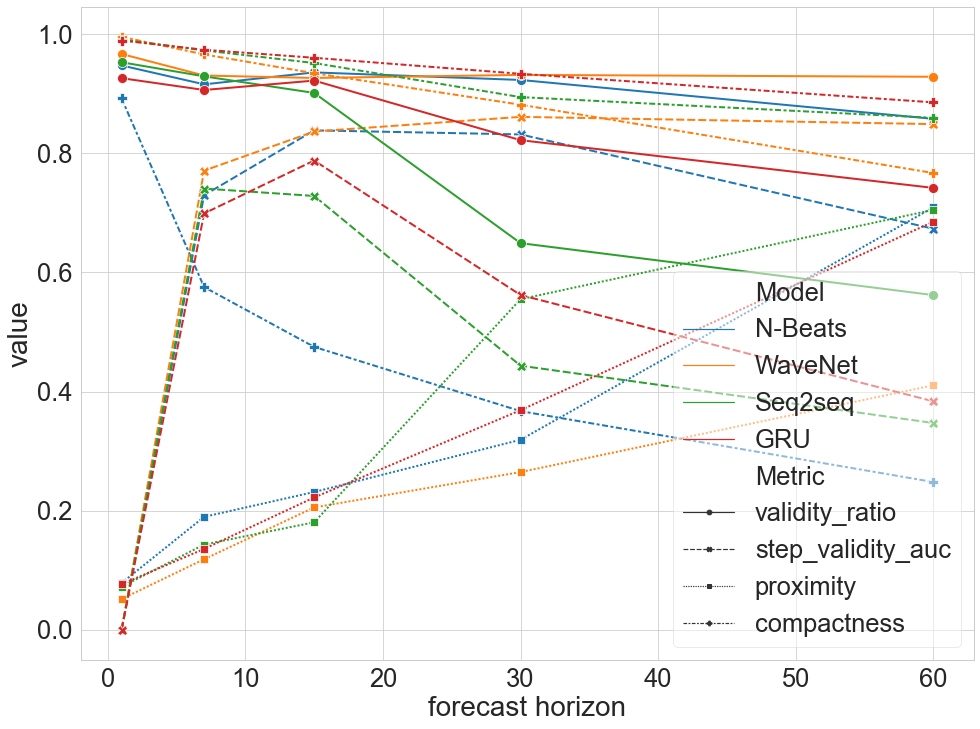}}
\hfill
\subfloat[MIMIC\label{subfig:1f}]{%
    \includegraphics[width=0.48\linewidth]{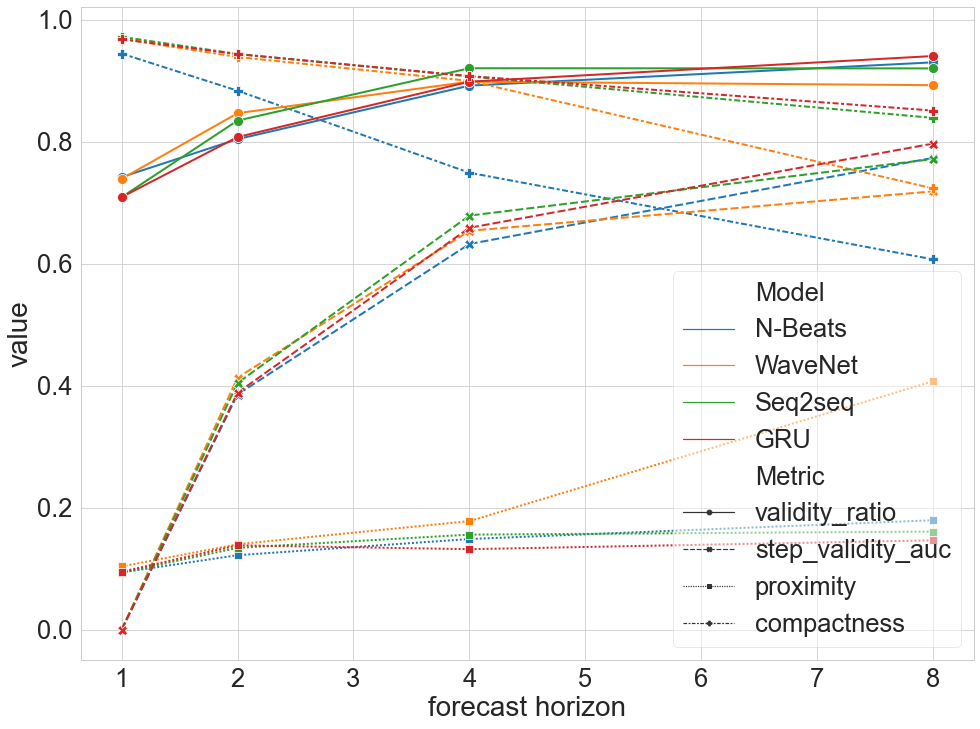}}
\caption{Horizon test by each dataset: each plot shows four evaluation metrics for four forecasting models, highlighted in different colours and shapes. The horizon increases gradually from $1$ to the defined horizon per dataset.}
\vspace{-4mm}
\label{fig:horizon-test}
\end{figure}

In the next experiment, we investigated the effectiveness of counterfactual generation in ForecastCF by gradually increasing the forecast horizon of DL models from $1$ to the defined horizon for each dataset. 
In Fig.~\ref{fig:horizon-test}, we observed an evident increasing trend for both \emph{validity ratio} and \emph{stepwise validity AUC} for the CIF2016 and MIMIC datasets (Fig.~\ref{subfig:1a} and Fig.~\ref{subfig:1f}); while for NN5, Tourism and M4 Finance, these metrics differed slightly, but most of them showed higher validity when increasing the horizon. Although for SP500, these two metrics decreased when the forecast horizon ranged from $15$ to $60$ (Fig.~\ref{subfig:1e}). This evidence suggests that when the forecast horizon of each DL model increases, the generated counterfactuals appear more valid in terms of the proportion of satisfying the desired ranges.
In addition, we found that the \emph{proximity} constantly increased and \emph{compactness} decreased for the majority of the datasets, especially for CIF2016, MIMIC and M4 Finance. In combination with the previous observation, we found a trade-off between \emph{validity} and \emph{data manifold closeness}: when the counterfactuals achieved higher validity ratio and stepwise AUC scores, more modifications were required for the original samples concerning the data manifold (i.e., less proximate and compact).

\begin{figure}[t]
\centering
\subfloat[Validity Ratio\label{subfig:2a}]{%
   \includegraphics[width=0.43\linewidth]{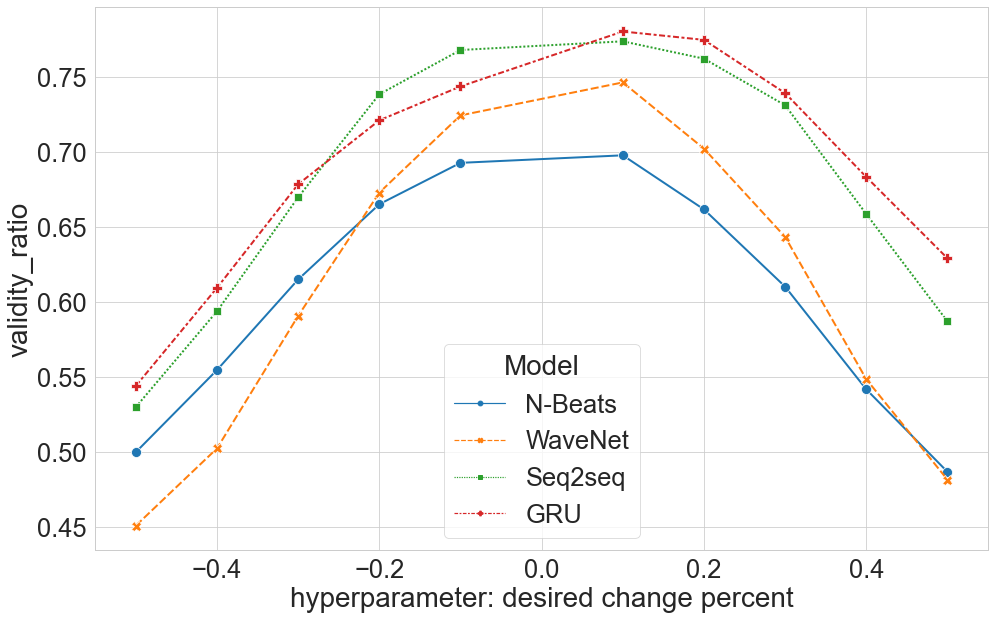}}
   \vspace{-2mm}
\hfill
\subfloat[Stepwise Validity AUC\label{subfig:2b}]{%
    \includegraphics[width=0.43\linewidth]{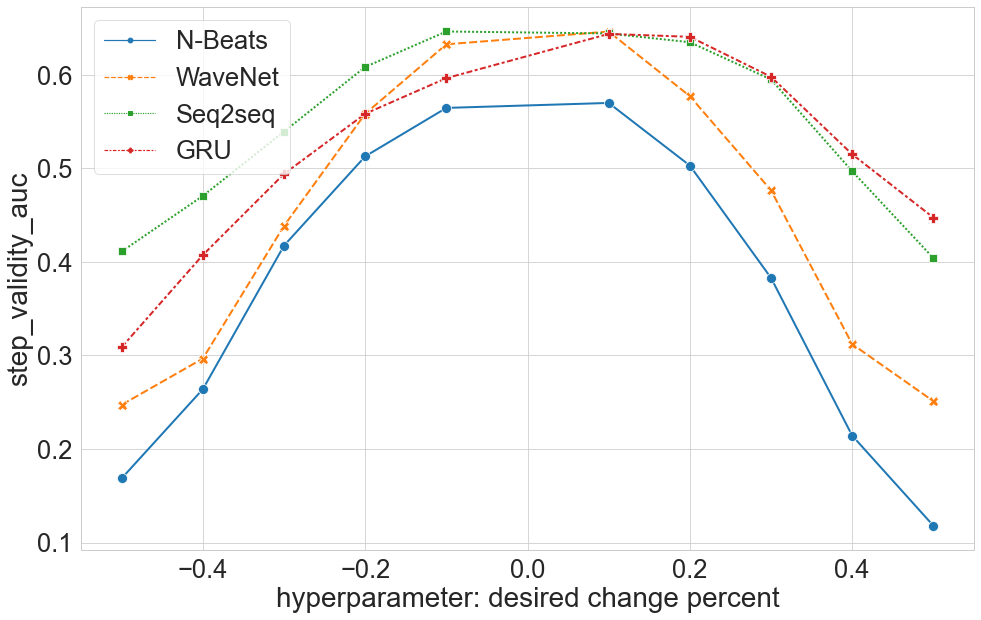}}
\\
\subfloat[Proximity\label{subfig:2c}]{%
    \includegraphics[width=0.43\linewidth]{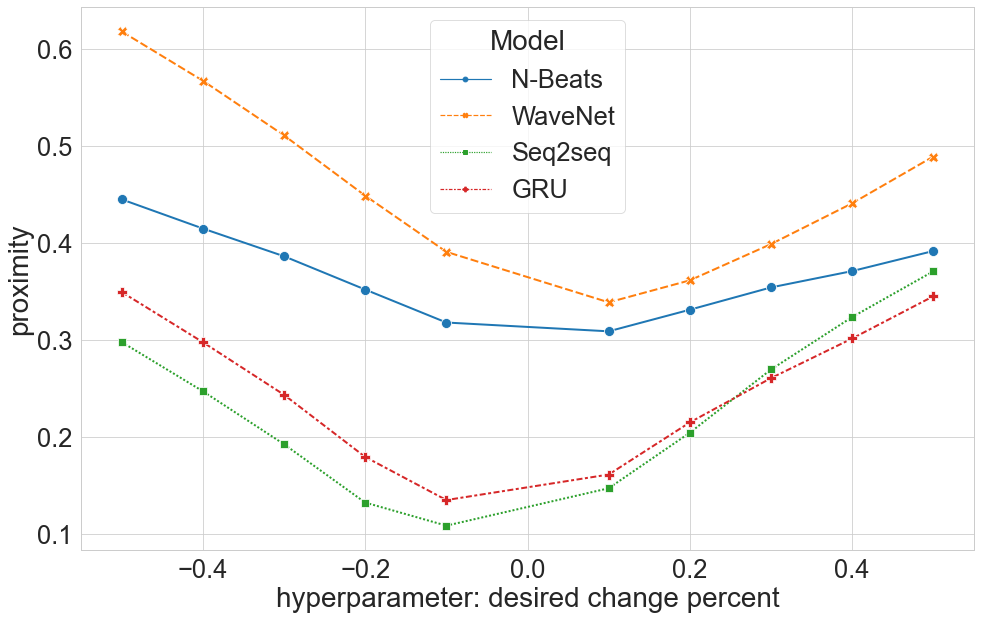}}
\hfill
\subfloat[Compactness\label{subfig:2d}]{%
    \includegraphics[width=0.43\linewidth]{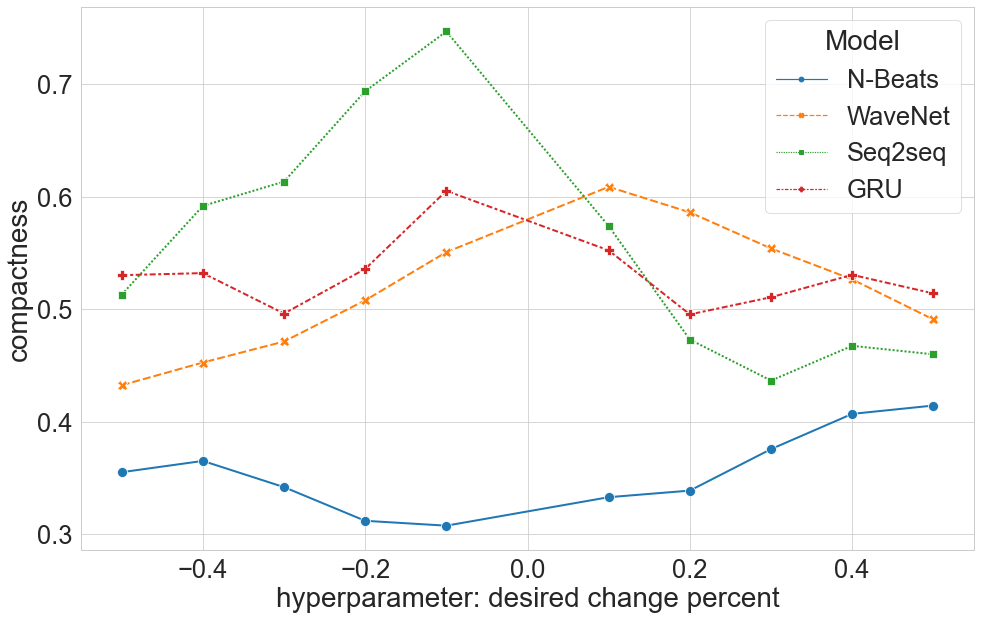}}
\caption{Ablation study: desired change percent $cp$.}
\label{fig:ablation-percent} 
\vspace{-4mm}
\end{figure}

\begin{figure}[t] 
\centering
\subfloat[Validity Ratio\label{subfig:3a}]{%
   \includegraphics[width=0.43\linewidth]{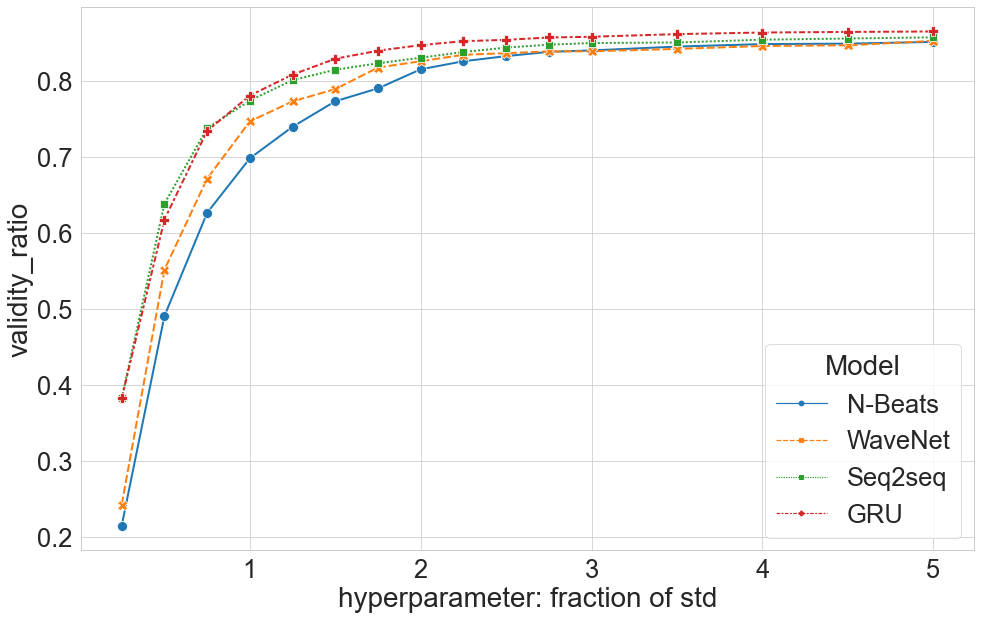}}
   \vspace{-2mm}
\hfill
\subfloat[Stepwise Validity AUC\label{subfig:3b}]{%
    \includegraphics[width=0.43\linewidth]{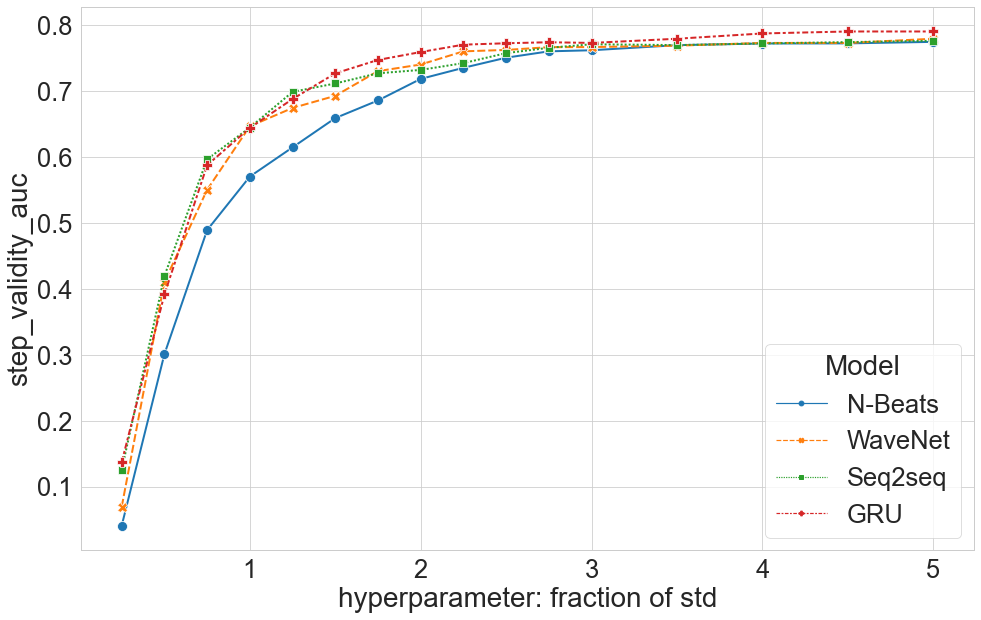}}
\\
\subfloat[Proximity\label{subfig:3c}]{%
    \includegraphics[width=0.43\linewidth]{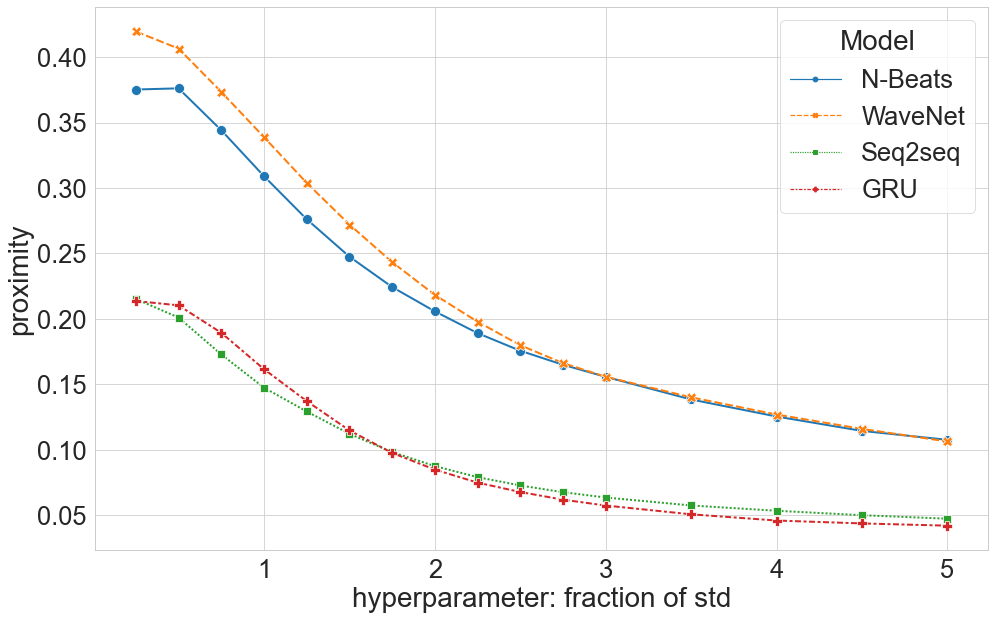}}
\hfill
\subfloat[Compactness\label{subfig:3d}]{%
    \includegraphics[width=0.43\linewidth]{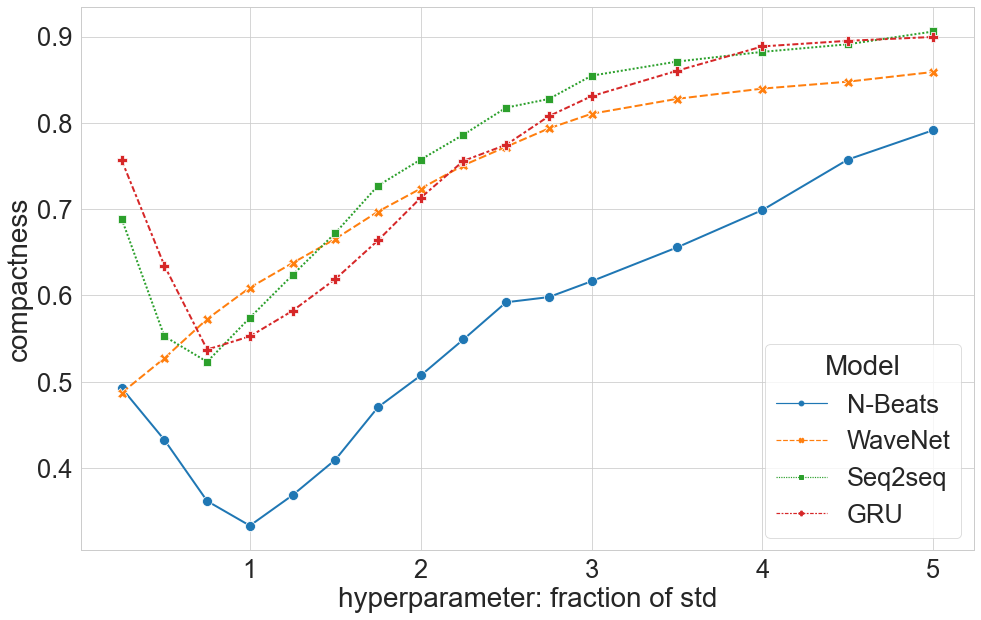}}
\caption{Ablation study: fraction of standard deviation $fr$.}
\label{fig:ablation-fraction} 
\vspace{-4mm}
\end{figure}

\smallskip
\noindent \textbf{Ablation study.} 
We performed an ablation study to examine the effects of two hyperparameters in ForecastCF: desired change percent $cp$ and fraction $fr$, 
while fixing the other hyperparameters as default. Both Fig.~\ref{fig:ablation-percent} and Fig.~\ref{fig:ablation-fraction} show the performance in terms of validity ratio, stepwise validity AUC, proximity and compactness on the \emph{CIF2016} dataset. 
We first observed that when $cp$ got closer to 0\% (i.e., closer to the vertical center in Fig.\ref{subfig:2a}-\ref{subfig:2d}), both the validity ratio and stepwise AUC became higher, while obtaining lower proximity and higher compactness. This finding suggests that when defining the desired prediction outcome, the minor percentage (compared to the original sample) we desire to change, the better the counterfactual performance. 
In Fig.\ref{subfig:3a}-\ref{subfig:3c}, we found that when $fr$ increased (i.e., with larger widths of the bounds), both validity ratio and stepwise AUC constantly increased while the proximity dropped, suggesting that defining larger constraint bounds could relieve the counterfactual condition and increase the model performance. However, we observed that compactness was the lowest when $fr$ was close to $1$, indicating that defining bounds that were too narrow could decline the performance. 
In sum, it is reasonable to perform an empirical search for the hyperparameters of ForecastCF to a new dataset to achieve more desirable performance. 

\begin{figure*}[ht] 
\centering
\subfloat[BaseNN\label{subfig:example1}]{%
   \includegraphics[width=0.33\linewidth]{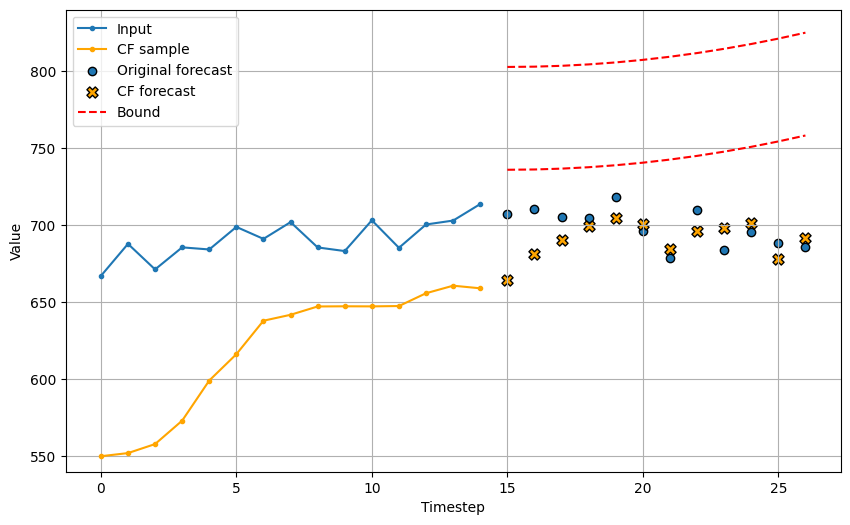}}
\hfill
\subfloat[BaseShift\label{subfig:example2}]{%
    \includegraphics[width=0.33\linewidth]{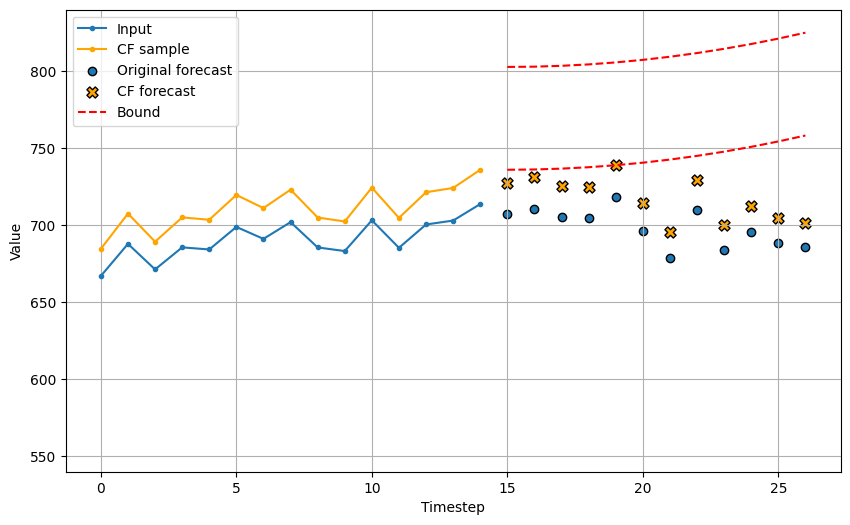}}
\hfill
\subfloat[ForecastCF\label{subfig:example3}]{%
   \includegraphics[width=0.33\linewidth]{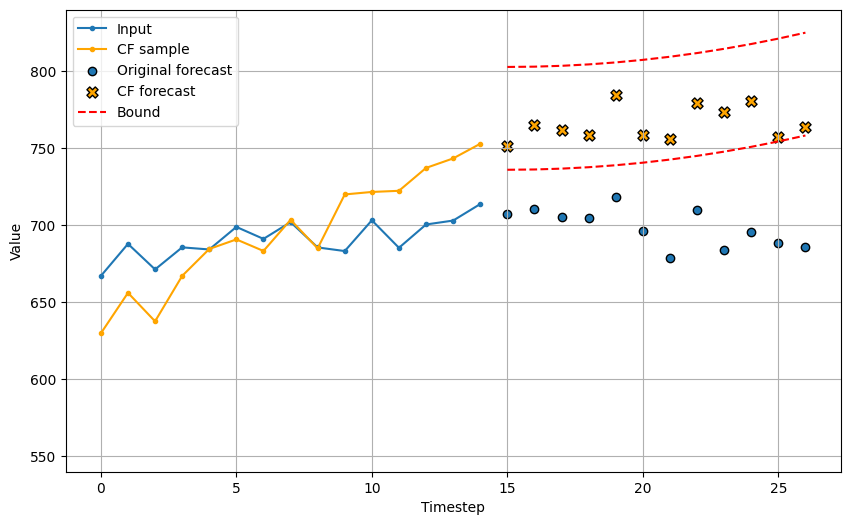}}
\\
\caption{Examples of generated counterfactuals from BaseNN, BaseShift and ForecastCF models using a time series example from the CIF2016 dataset.
The input time series is shown as the blue line, and the counterfactual sample is the yellow line. The red-dotted curves define the forecast constraints that the predicted values (yellow points) of the counterfactuals are more desired to fall into; the blue points denote the forecasted values on the original sample.}
\label{fig:example-results}
\vspace{-2mm}
\end{figure*}

\smallskip
\noindent \textbf{Examples.}
Fig.~\ref{fig:example-results} provides three individual counterfactuals from CIF2016 as a qualitative analysis of the performance difference between ForecastCF and the baseline models. 
In Fig.~\ref{subfig:example3}, we observed that ForecastCF had the most proximate and compact counterfactual (the yellow line) regarding the original time series (the blue line) compared to the other two baselines, while the counterfactual from BaseNN diverged significantly from the original sample (Fig.~\ref{subfig:example1}). 
In terms of validity, we found that all the $12$ predicted values (yellow points) of the counterfactual from ForecastCF fell within the desired forecasting trend (red-dotted lines), hence this was a fully valid counterfactual. On the other hand, we observed that both BaseNN and BaseShift had no valid predicted values (yellow points) in Fig.~\ref{subfig:example1} and \ref{subfig:example2}. 
These findings are aligned with our experimental results and they further demonstrate that ForecastCF could generate more relevant and valid counterfactuals for forecasting.

\smallskip
\noindent \textbf{Runtime analysis.}
We additionally conducted an analysis of the runtime for different counterfactual models using 50 test samples from the CIF2016 dataset (Table~\ref{tab:runtime-analysis}). We found that BaseNN was the most efficient in terms of counterfactual generation due to the nature of the method, while ForecastCF for the WaveNet model consumed the most computational time (i.e., 986 seconds for 50 samples) as a complex DL model. We compared the four different forecasting models with ForecastCF and found that GRU was the most efficient among them, suggesting that simpler DL models would take less computational time for counterfactual generation. 

\begin{table}[tb]
\caption{Runtime (in sec.) for generated counterfactual samples in CIF2016.}\label{tab:runtime-analysis}
\vspace{-4mm}
\begin{center}
\begin{tabular}{c c c c|c c }
\hline
\multicolumn{4}{c|}{\textbf{ForecastCF}} &\multicolumn{2}{|c}{\textbf{Baseline}} \\
\textit{N-Beats}& \textit{WaveNet} & \textit{Seq2seq}& \textit{GRU} & \textit{BaseNN}& \textit{BaseShift}  \\
\hline
156.5315  & 986.8895 & 162.8258 & 54.3887 & 0.0001 & 0.0032\\
\hline
\end{tabular}
\vspace{-6mm}
\end{center}
\end{table}

\section{Conclusions}
We formulated the problem of counterfactual explanations for time series forecasting, and proposed a gradient-based solution, ForecastCF, with different instantiations of defining the desired prediction outcome.
Our experimental results with four deep forecasting architectures showed that ForecastCF outperformed two baseline models in terms of quantitative metrics that measure counterfactual validity and data manifold closeness. 
ForecastCF additionally demonstrated good trade-offs between validity and data manifold closeness under the constraint of desired forecast ranges. 
Future work involves extending our solution into other forecasting models, such as traditional statistical models that do not rely on DL architectures. 
Furthermore, we intend to investigate applying ForecastCF in a multivariate forecasting setup, and to integrate exogenous variables in explaining the forecasting models.  
For reproducibility, the source code is publicly available on our supporting website\footnote[5]{\url{https://github.com/zhendong3wang/counterfactual-explanations-for-forecasting}}.

\bibliographystyle{IEEEtran}
\bibliography{main}

\end{document}